\documentclass[pdflatex,sn-mathphys-num]{sn-jnl}

\usepackage[T1]{fontenc}
\usepackage{graphicx}
\usepackage{multirow}
\usepackage{multicol}
\usepackage{makecell}
\usepackage{textcomp}
\usepackage{xcolor}
\usepackage{tabularx}
\usepackage[ruled,vlined,linesnumbered]{algorithm2e}
\usepackage{booktabs}
\usepackage{threeparttable}
\usepackage{enumitem}
\usepackage{listings}
\usepackage{array}
\usepackage{rotating}
\usepackage{pdflscape}
\usepackage{amsmath,amssymb,amsfonts}
\usepackage{amsthm}
\usepackage{ifthen}

\newcolumntype{P}[1]{>{\centering\arraybackslash}p{#1}}

\theoremstyle{definition}
\newtheorem{defn}{Definition}

\begin{document}

\newcommand{\definenb}[2]{%
  \expandafter\newcommand\csname #1\endcsname[1]{\nb{#2}{##1}}%
}

\newboolean{showcomments}
\setboolean{showcomments}{false} 
\ifthenelse{\boolean{showcomments}}
{\newcommand{\nb}[2]{
		\fcolorbox{black}{yellow}{\bfseries\sffamily\scriptsize#1}
		{\sf\small$\blacktriangleright$\textit{#2}$\blacktriangleleft$}
	}
	\newcommand{\version}{\emph{\scriptsize$-$working$-$}}
}
{\newcommand{\nb}[2]{}
	\newcommand{\version}{}
}

\newcommand\jialong[1]{\nb{Jialong}{#1}}
\newcommand\kenji[1]{\nb{Kenji}{#1}}
\newcommand\enhong[1]{\nb{Enhong}{#1}}

\newcommand{\methodname}{\textit{SMART}}
\renewcommand\cellalign{tr}
\setlength{\tabcolsep}{2pt}
\renewcommand{\arraystretch}{0.95}
\title[Coverage-Aware Game Playtesting]{Synergizing Code Coverage and Gameplay Intent: Coverage-Aware Game Playtesting with LLM-Guided Reinforcement Learning}

\author[1]{\fnm{Enhong} \sur{Mu}}\email{muenhong@email.swu.edu.cn}
\author[2]{\fnm{Minami} \sur{Yoda}}\email{yoda.minami@nihon-u.ac.jp}
\author[3]{\fnm{Yan} \sur{Zhang}}\email{zhangyan@sz.tsinghua.edu.cn}
\author[1]{\fnm{Mingyue} \sur{Zhang}}\email{myzhangswu@swu.edu.cn}
\author[2]{\fnm{Yutaka} \sur{Matsuno}}\email{matsuno.yutaka@nihon-u.ac.jp}
\author[4]{\fnm{Jialong} \sur{Li}}\email{lijialong@fuji.waseda.jp}

\affil[1]{\orgname{College of Computer and Information Science, Southwest University}, \orgaddress{\city{Chongqing}, \country{China}}}
\affil[2]{\orgname{College of Science and Technology, Nihon University}, \orgaddress{\city{Chiba}, \country{Japan}}}
\affil[3]{\orgname{Shenzhen International Graduate School, Tsinghua University}, \orgaddress{\city{Shenzhen}, \country{China}}}
\affil[4]{\orgname{Waseda Institute for Advanced Study, Waseda University}, \orgaddress{\city{Tokyo}, \country{Japan}}}








            
            
            


\abstract{%
The widespread adoption of the ``Games as a Service'' model necessitates frequent content updates, placing immense pressure on quality assurance. In response, automated game testing has been viewed as a promising solution to cope with this demanding release cadence. However, existing automated testing approaches typically create a dichotomy: code-centric methods focus on structural coverage without understanding gameplay context, while player-centric agents validate high-level intent but often fail to cover specific underlying code changes. To bridge this gap, we propose \methodname{} (\textbf{S}tructural \textbf{M}apping for \textbf{A}ugmented \textbf{R}einforcement \textbf{T}esting), a novel framework that synergizes structural verification and functional validation for game update testing. \methodname{} leverages Large Language Models (LLMs) to interpret Abstract Syntax Tree (AST) differences and extract functional intent, constructing a context-aware hybrid reward mechanism. This mechanism guides Reinforcement Learning agents to sequentially fulfill gameplay goals while adaptively exploring modified code branches. We evaluate \methodname{} on two environments, \textit{Overcooked} and \textit{Minecraft}. The results demonstrate that \methodname{} significantly outperforms state-of-the-art baselines; it achieves over 94\% branch coverage of modified code, nearly double that of traditional RL methods, while maintaining a 98\% task completion rate, effectively balancing structural comprehensiveness with functional correctness.
}




\keywords{Game Testing, Game Updates, Large Language Model, Reinforcement Learning}

\maketitle

\section{Introduction}
The global video game industry has evolved into an economic and cultural powerhouse, with market revenues reaching hundreds of billions of dollars annually~\cite{newzoo2024}. In such a fiercely competitive landscape, the software quality assurance (QA) is not merely a technical step but a cornerstone of commercial success. Rigorous game testing is essential to ensure functionality, stability, and a polished user experience, as post-launch bugs can severely damage player trust, brand reputation, and revenue streams~\cite{RLReg}. 

This landscape is further complicated by the industry's widespread adoption of the ``Games as a Service'' (GaaS) model. Unlike traditional single-purchase titles, GaaS products are continuously evolving entities, sustained by a frequent cadence of updates and content patches (e.g., new game quests and materials) designed to maintain player engagement~\cite{Politowski2021A}. 
While these updates are crucial for player retention and monetization, each patch---whether introducing new features or modifying existing ones---carries the inherent risk of introducing unforeseen defects into a stable codebase. This relentless development cycle places immense pressure on testing teams, demanding automated testing methodologies that are highly efficient and responsive to rapid iteration.

In response to these demands, automated game testing has emerged as a critical area of research. These studies can be bifurcated into two distinct paradigms. The first paradigm, rooted in conventional software engineering, adopts a code-centric perspective. Utilizing white-box or grey-box techniques, it prioritizes the structural testing of the program through methods such as unit testing and static analysis. The principal metric in this line of work is code coverage, aiming to ensure that every line of logic within the codebase is executed.
In contrast, the second paradigm is player-centric, employing beta testing to focus on the functional correctness and behavioral validity of the game. Studies in this tradition validate high-level gameplay scenarios---such as whether a quest can be successfully completed. The overarching goal is to ensure that the game functions as intended from an end-user perspective.

However, we argue that this dichotomous division is increasingly insufficient for the modern "Game as a Service" (GaaS) model, where games are subject to content updates. A typical game content update inherently constitutes a hybrid of functional and structural changes: it may introduce new quests (macro-level functional gameplay intent) while also incorporating new elements (such as new items, new interaction methods, etc.) (micro-level structural changes). 
As such, single-dimensional testing approaches are inadequate. For example, a test suite focused solely on code coverage may confirm the execution of a modified numerical parameter but cannot determine whether the adjustment yields the desired gameplay intent. Conversely, functional tests that only assess new quest flows may entirely fail to cover all the new code.

To this end, this paper proposes \methodname (\textbf{S}tructural \textbf{M}apping for \textbf{A}ugmented \textbf{R}einforcement \textbf{T}esting), an incremental hybrid testing framework designed to systematically integrate both structural verification and functional validation. Specifically, we employ Large Language Models (LLMs) to interpret code updates, decomposing them into sequential functional subgoals aligned with specific structural anchors. These inputs are integrated into an adaptive hybrid reward mechanism that motivates the Reinforcement Learning (RL) agent to pursue dual objectives: validating gameplay intent through goal completion, while exploring edge-case interactions to maximize the coverage of modified code branches.

The contributions of this paper are as follows:
\begin{itemize}
    \item We propose \methodname{}, a novel testing framework that systematically combines low-level code modifications with high-level functional validation. Specifically, \methodname{} utilizes a hybrid reward mechanism--integrating LLM---extracted semantic signals with AST-based structural signals---to guide agents in validating gameplay logic while actively exploring edge-case interactions within the modified code.
    
    \item We introduce a context-aware mapping methodology to enable precise hybrid guidance. The functional intent of an update is decomposed into ordered \textit{semantic subgoals}, while low-level code modifications are represented as \textit{structural anchors}. A dynamic mapping between the two ensures that the agent is rewarded only for covering structural changes relevant to the current gameplay stage, thereby preventing inefficient exploration of irrelevant or unreachable code paths.
    
    \item We conduct experiments on two games, Overcooked and Minecraft, to evaluate the effectiveness of \methodname{}. 
\end{itemize}

The rest of this paper is structured as follows. Section~\ref{sec:background} establishes the technical background. 
Section~\ref{sec:proposal} elaborates on the design and implementation of our \methodname{} framework. Section~\ref{sec:evaluation} presents our experimental setup, results, and discussion. 
Section~\ref{sec:relatedwork} reviews prior work on automated game testing and the application of LLMs. 
Finally, Section~\ref{sec:conclusion} concludes the paper and outlines future work.

\section{Background}
\label{sec:background}
\subsection{Abstract Syntax Tree (AST)}

An Abstract Syntax Tree (AST) is a tree-based representation of the syntactic structure of source code \cite{aho2006compilers}. From the perspective of graph theory, an AST can be regarded as a special kind of Directed Acyclic Graph (DAG), where each node represents a syntactic construct in the source code, and edges represent the hierarchical and logical relationships between these constructs.

Formally, a family of ASTs, denoted $\mathcal{A}[\mathcal{X}]$, can be defined inductively. Let $S$ be a finite set of sorts (for example, ``expression'', ``statement'', etc.), and let $\mathcal{O}$ be a set of operators (such as `+', `if'), each associated with an arity describing the types of its arguments. Given a family of variable sets $\mathcal{X}$ indexed by sorts, the family of ASTs $\mathcal{A}[\mathcal{X}]$ is the smallest collection satisfying:
\begin{itemize}
    \item Variables are ASTs: If $x \in \mathcal{X}_s$, then $x \in \mathcal{A}[\mathcal{X}]_s$.
    \item Operators combine ASTs: If an operator $o$ has arity $(s_1, \dots, s_n)s$, and for all $1 \leq i \leq n$, $a_i \in \mathcal{A}[\mathcal{X}]_{s_i}$, then $o(a_1, \dots, a_n) \in \mathcal{A}[\mathcal{X}]_s$.
\end{itemize}

The ``abstract'' in AST distinguishes it from the Concrete Syntax Tree (CST, or parse tree) that is typically generated by the parser in the early stages of compilation. While a concrete syntax tree faithfully represents every syntactic detail from the source text, the AST omits non-essential elements for structural understanding, such as parentheses used for grouping, semicolons terminating statements, and formatting artifacts like indentation or newlines. Instead, the AST is designed to capture the structural and semantic content of the code, rather than its surface textual representation.

Within the AST, different language constructs are clearly expressed through their hierarchical relationships. For example, an \textit{if-else statement} is usually represented by an `If' node, which has at least two children: the first child denotes the condition expression, and the second child represents the `then' code block. If there is an `else' or `else if' branch, it appears as a third child (the \textit{orelse} node). Another example is the \textit{loop statement}: a `for' loop is typically represented by a `For' node, which has multiple children corresponding to the initialization variable, the loop termination condition, the update operation after each iteration, and the loop body itself.
Thanks to this hierarchical tree structure, the AST is typically a direct product of the syntactic analysis phase and serves as a crucial intermediate representation (IR) throughout many stages of software compilation and analysis \cite{738528,10.5555/1251398.1251416}.

\subsection{Reinforcement Learning}

Reinforcement Learning (RL) is an important branch of machine learning that studies how an agent can learn to make decisions through interactions with an environment, with the objective of maximizing the cumulative reward it receives \cite{sutton2018reinforcement}. This learning paradigm is particularly well-suited to solving complex problems that require sequential decision-making, such as video games \cite{mnih2015human}.

The RL problem is typically modeled as a Markov Decision Process (MDP), which is formally defined by a five-tuple $(S, A, P, R, \gamma)$. Here, $S$ denotes the set of all possible environment states, while $A$ represents the set of actions that the agent can perform. The transition probability function $P(s'|s, a)$ defines the probability of transitioning to state $s'$ when the agent takes action $a$ in state $s$. The reward function $R(s, a, s')$ specifies the immediate reward signal the agent receives after performing action $a$ in state $s$ and transitioning to $s'$. The discount factor $\gamma \in [0, 1]$ balances the importance of short-term and long-term rewards.
The agent's behavior is determined by its policy $\pi(a|s)$, which gives the probability of selecting action $a$ when in state $s$. The goal of RL algorithms is to learn an optimal policy $\pi^*$ that maximizes the expected cumulative discounted reward—also known as return—from any initial state.

Among the many RL algorithms, Proximal Policy Optimization (PPO) has become one of the most popular and effective approaches due to its excellent sample efficiency and training stability \cite{schulman2017proximalpolicyoptimizationalgorithms}. PPO belongs to the family of policy gradient methods and optimizes a ``clipped'' surrogate objective to restrict the step size of each policy update. The core objective function is defined as follows:
\begin{equation}
L^{CLIP}(\theta) = \hat{\mathbb{E}}_t \left[ \min \left( r_t(\theta) \hat{A}_t, \mathrm{clip}(r_t(\theta), 1 - \epsilon, 1 + \epsilon) \hat{A}_t \right) \right]
\end{equation}
In this formulation, $\hat{\mathbb{E}}_t$ denotes the empirical average over timesteps $t$. The term $r_t(\theta) = \frac{\pi_{\theta}(a_t|s_t)}{\pi_{\theta_{\text{old}}}(a_t|s_t)}$ represents the probability ratio of the new and old policies for a given action, where $\pi_{\theta}$ is the current policy being optimized and $\pi_{\theta_{\text{old}}}$ is the policy from the previous iteration. The term $\hat{A}_t$ denotes an estimate of the advantage function at time $t$, which measures how much better (or worse) taking action $a_t$ in state $s_t$ is compared to the average. The function $\mathrm{clip}(x, \min, \max)$ restricts its input $x$ to the interval $[\min, \max]$, and $\epsilon$ is a small hyperparameter (typically 0.2) defining the clipping range.
By constraining the probability ratio $r_t(\theta)$ within a certain range, this objective ensures that policy updates do not deviate too far from the previous policy, thereby improving the stability of the learning process.

\section{Proposal}
\label{sec:proposal}

\subsection{Overview}
\begin{figure*}[htb]
\centering
\includegraphics[width=0.96\linewidth]{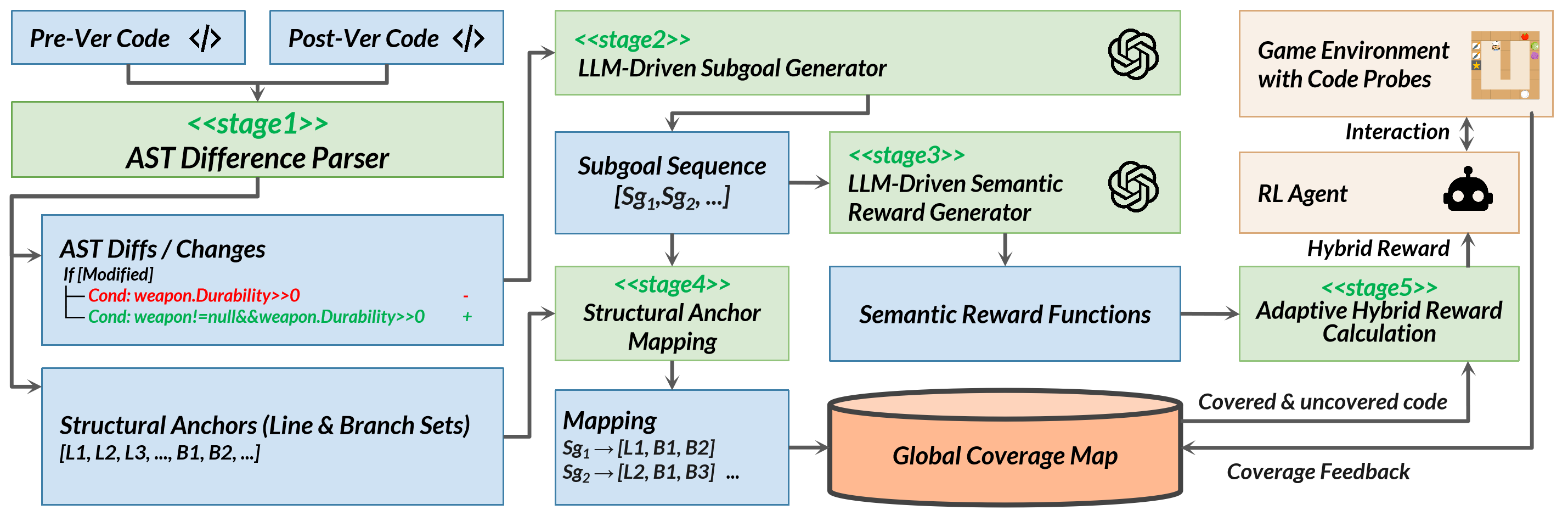}
\caption{
Architectural overview of the \methodname{} framework, where green boxes represent the core processing stages of the framework, blue boxes denote the intermediate data artifacts.
}
\label{fig:framework}
\end{figure*}

As illustrated in Figure~\ref{fig:framework}, \methodname{} synergizes the structural code changes and functional gameplay intent through a systematic five-stage pipeline. The pipeline consists of five stages: (i) AST Difference Parsing, which identifies all modified lines and branches between game versions by comparing their ASTs; (ii) Semantic Subgoal Generation, which decomposes the new quests into an ordered sequence of natural language subgoals using LLMs; (iii) Semantic Reward Generation, which converts the natural language subgoal sequence into a corresponding sequence of machine-executable RL reward functions; (iv) Structural Anchor Mapping, which establishes a context-aware link between each low-level code and branch change and the specific high-level subgoal(s) it is related to; and (v) Adaptive Hybrid Reward Function, which conducts the testing process by guiding an RL agent with a hybrid reward that combines sequential subgoal completion with the discovery of novel code coverage. This hybrid reward function effectively separates concerns: the semantic reward ensures the agent learns \textit{what} to do (functional correctness), while the adaptive structural reward ensures it explores \textit{how else} it can be done (structural comprehensiveness).

In the remainder of this section, we first provide a motivating example, followed by detailed technical descriptions of each component.

\subsection{Motivating Example}
\label{sec:motivating_example}



To more clearly and intuitively illustrate our approach, we present a motivating example from an Overcooked-style cooking game. Suppose a game update adds items such as dough and an oven, which then creates several quests. One of these quests is called "Onion Pizza". This quest requires the player to prepare a onion pizza by assembling multiple ingredients in a specific order and baking the assembled pizza in an oven.

Concretely, the updated quest logic requires the player to (i) obtain a dough base and place it on a counter, (ii) fetch a tomato and chop it on a cutting board, (iii) add chopped tomato and cheese onto the dough to form a raw tomato pizza, (iv) bake the raw pizza in the oven until it becomes a finished pizza, (v) place the baked pizza on a plate, and then (vi) fetch an onion, chop it, and sprinkle the chopped onion onto the plated pizza before finally delivering it at the delivery station. This process forms a multi-step cooking sequence that naturally decomposes into a set of intermediate gameplay subgoals.

At the same time, the Pizza Onion quest update introduces several low-level modifications in the codebase. These include the definition of new item states for tomatoes and onions (e.g., whole vs. chopped), the handling of dough, tomato, and cheese combinations on counters, the baking logic inside the oven that transforms a raw pizza into a finished pizza, and the post-baking logic that allows a plated pizza to accept an additional onion topping. Additional code is also added around the delivery station to validate that only correctly assembled and decorated onion pizzas are accepted as valid quest completions.

\subsection{AST Difference Parser}
After a game update, the AST Difference Parser first compares the source code of the new and old versions to identify the locations and content of code changes. Specifically, it constructs ASTs for both the pre-update and post-update versions of the code. A tree-differencing algorithm then compares these two ASTs to identify added, modified, and deleted nodes, ignoring purely stylistic changes like whitespace.

The raw output of this process is a detailed list of AST node differences. To make these changes actionable for testing, this stage processes the raw diffs to populate two distinct sets of testable sets, ($C_L$, $C_B$), which we term structural anchors:
\begin{itemize}[leftmargin=*]
    \item \textbf{Modified Lines ($C_L$)}: This is the set of individual lines of code that have been added or substantively modified. Each entry $l \in C_L$ is uniquely identified by its location (e.g., file path and line number) and represents a discrete statement that must be executed to be considered covered.
    \item \textbf{Modified Branches ($C_B$)}: This is the set of newly introduced or altered logical branches within control-flow statements (e.g., `if', `switch', `while'). A change to a control-flow node in the AST—such as modifying the condition of an `if' statement—results in its corresponding logical paths being added to this set. For instance, a modified `if-else' statement contributes two branches to $C_B$: one for the `true' condition and one for the `false' condition. Each branch $b \in C_B$ is identified by the location of its control statement and the specific condition that activates it.
\end{itemize}



\subsection{Stage 2: Subgoal Generation}
Given the code updating the corresponding to one new game quest, the LLM-powered \textit{Subgoal Generator} is tasked with analyzing the raw code changes associated with a single, complex game quest and decomposing it into a logically ordered sequence of subgoals. By analyzing function calls, data dependencies (e.g., a function requiring `chopped\_tomato' as input), and quest logic extracted from the AST Diffs, the LLM infers the necessary intermediate steps. 
For instance, after parsing the code updates related to the "Onion Pizza" task, the Subgoal generator sequentially identifies a series of verifiable natural language sub-goals from the overall task, such as "get and place the dough", "process the tomatoes and assemble the raw pizza", and "bake the pizza and add chopped onions", forming an ordered sequence of sub-goals (see Listing~\ref{lst:subgoal_example}).

The final output of this stage is a semantically meaningful and ordered sequence of subgoals, denoted as $S = (sg_1, sg_2, \dots, sg_n)$. Each element $sg_j \in S$ is a natural language string describing a distinct, verifiable step of the overall task. For the pizza quest example, this sequence would be: (i) $sg_1$: "Obtain and chop a tomato"; (ii) $sg_2$: "Process the chopped tomato into sauce using a mixer"; (iii) $sg_3$: "Assemble the pizza on a base with sauce and cheese"; and (iv) $sg_4$: "Bake the assembled pizza in the oven."

\subsection{Stage 3: Semantic Reward Generation}
To make the abstract subgoals from Stage 2 actionable, the LLM-driven semantic reward generator translates each step's objective into a formally defined, executable reward rule. This component is provided with the subgoal sequence $S$ and the game's environment observation schema, a manifest detailing all observable state variables.

For each subgoal $sg_j$ in the sequence $S$, the LLM is prompted to generate a specific reward rule. This rule defines a success condition for that particular step using only the variables available in the observation schema. This process transforms the high-level natural language plan into a machine-verifiable format.
For instance, for the sub-goal "baking pizza", the generator creates one reward for putting the raw pizza into the oven; one reward for each observable improvement in baking progress; and one final reward triggered when the pizza transitions from a raw state to a finished state, then outputs them in JSON format.(see Listing~\ref{lst:reward_example})

Based on the subgoal sequence $S = (sg_1, sg_2, \dots, sg_n)$ from Stage 2, this stage generates a corresponding ordered sequence of reward functions, denoted as $R = (r_1, r_2, \dots, r_n)$. Each function $r_j: \text{State} \to \mathbb{R}$ is a binary predicate corresponding to subgoal $sg_j$. It returns a positive reward if and only if the success conditions for $sg_j$ are met in the current game state, and zero otherwise. This sequence $R$ provides the series of checkpoints that the agent must achieve in the correct order, forming the backbone of the guidance mechanism detailed in Stage 5.

\subsection{Stage 4: Structural Anchor Mapping}
This stage forges the critical link between the high-level subgoals and the specific low-level code that needs to be triggered. In a multi-step quest, the code relevant to the player changes as they progress. For instance, after a tomato is chopped (Subgoal 1), the code related to the interaction between a whole tomato and a cutting board will never again be triggered during normal gameplay. Therefore, a context-aware mapping—i.e., determining which code segments should be tested at which subgoal stage—is essential.

\begin{defn}[Structural Anchor Mapping]
\label{def:anchor_mapping}
Let $S = $ \\ $\{sg_1, sg_2, \dots, sg_n\}$ be the ordered set of subgoals from Stage 2. Let $C_L$ and $C_B$ be the sets of modified lines and branches, respectively, identified by the AST Difference Parser in Stage 1. We define the comprehensive set of all structural anchors as $C = C_L \cup C_B$. The goal of this stage is to construct a mapping function $M: C \to \mathcal{P}(S)$, where $\mathcal{P}(S)$ is the power set of $S$. This function assigns each structural anchor $c_i \in C$ to a non-empty subset of subgoals $M(c_i) \subseteq S$ during which it is considered relevant and testable. The mapping must be comprehensive, ensuring that for every anchor $c_i$, its assigned set of subgoals $M(c_i)$ is not empty. A single anchor can be mapped to multiple subgoals (i.e., $|M(c_i)| \ge 1$).
\end{defn}

To construct this mapping, we employ a two-phase hybrid approach.  
First, we perform a static analysis to establish a coarse-grained reachability relationship. For each subgoal, we identify its primary entry point in the code. From these entry points, we trace backwards through the call graph and data dependencies to identify a candidate set of all potentially reachable structural anchors from $C$. This initial pass provides a conservative over-approximation.  
The candidate sets from Phase 1, while safe, may contain anchors that are technically reachable but not semantically relevant. Therefore, in the second phase, the LLM acts as a sophisticated filter. For each subgoal $sg_j$ and each structural anchor $c_i$ in its candidate set, we prompt the LLM with the subgoal’s description and the anchor’s code. The LLM then determines whether the anchor is \textit{semantically and logically} pertinent to achieving that specific subgoal, and prunes the mapping if it is not. 

For instance, in the Pizza Onion quest, a version update may have modified (i) the logic for checking whether a dough on a counter already holds a valid tomato and cheese combination and (ii) refined the state transitions for chopped tomatoes and onions. These structural anchors are only relevant to specific subgoals such as “assemble a raw tomato pizza” and “bake the pizza in the oven,” and not to earlier steps like ``place the dough on a counter.”
Without structural-anchor mapping, the agents may repeatedly interact with the oven or attempt to place onions on plates before any dough or baked pizza is available. By contrast, our two-phase mapping can accurately identify (i) that dough-combination logic corresponds to the assembly subgoal and (ii) that oven-related logic corresponds to the baking subgoal in the Pizza Onion pipeline. (see Listing~\ref{lst:mapping_example})

\subsection{Stage 5: Adaptive Hybrid Reward Function}
\label{sec:stage5}
The final stage is the adaptive hybrid reward function that orchestrates the agent's learning process. Its primary goal is to train an agent to successfully execute the entire sequence of subgoals within a single episode, while simultaneously encouraging comprehensive exploration of all modified code across multiple episodes.

To achieve this, the function logic, detailed in Algorithm~\ref{alg:training_loop_algorithm2e}, tracks two levels of state: an in-episode progress tracker ($j$) that resets every episode, and a global coverage map ($C_{cov}$) that persists throughout the entire training process. At each step, it calculates a hybrid reward by checking for sequential subgoal completion against the reward sequence $R$ and for novel code coverage against the global map.

\begin{algorithm}[htb]
\SetAlgoLined 
\DontPrintSemicolon 

\caption{Adaptive Hybrid Reward Training Loop}
\label{alg:training_loop_algorithm2e}

\KwIn{
    Subgoal sequence $S = (sg_1, \dots, sg_n)$\;
    Reward functions $R = (r_1, \dots, r_n)$\;
    Structural anchors $C$\;
    Reward magnitudes $r^{sem}, r^{str}$\;
}

Initialize global coverage map $C_{cov} \leftarrow \emptyset$\;

\BlankLine 

\For{each episode $e = 1, \dots, E$}{
    $s_0 \leftarrow \text{env.reset()}$ \tcp*{Reset environment}
    $j \leftarrow 1$ \tcp*{Reset in-episode subgoal index}
    $\mathcal{D}_e \leftarrow \emptyset$ \tcp*{Initialize trajectory buffer}
    
    \For{each step $t = 0, \dots, T-1$}{
        $a_t \leftarrow \pi(s_t)$ \tcp*{Agent selects an action}
        $(s_{t+1}, d_t, C_{step}) \leftarrow \text{env.step}(a_t)$
        
        \tcp{\textbf{Hybrid Reward Calculation}}
        $r_t \leftarrow 0$\;
        
        \If{$j \le n$ \textbf{and} $r_j(s_{t+1}) > 0$}{
            $r_t \leftarrow r_t + r^{sem}$ \tcp*{Semantic reward}
            $j \leftarrow j + 1$ \tcp*{Advance to next subgoal}
        }
        
        \ForEach{anchor $c$ in $C_{step}$}{
            \If{$c \notin C_{cov}$}{
                $r_t \leftarrow r_t + r^{str}$ \tcp*{Structural reward}
                $C_{cov} \leftarrow C_{cov} \cup \{c\}$ \tcp*{Mark covered anchor}
            }
        }
        
        Store $(s_t, a_t, r_t, s_{t+1}, d_t)$ in $\mathcal{D}_e$\;
        \If{$d_t$}{
            \textbf{break}\;
        }
    }
        Update policy $\pi$ using transitions in $\mathcal{D}_e$;
}
\end{algorithm}

To illustrate the algorithm's dynamics, consider the pizza-making task. In an early episode, the agent's primary motivation is to earn the sequential semantic rewards. It might learn to pick up a tomato and use the cutting board, which satisfies $sg_1$ ("Obtain and chop a tomato"). Upon completion, the check at line 10 becomes true, the agent receives a large reward $r^{sem}$, and its internal target advances to $sg_2$ (the subgoal index $j$ is incremented to 2). During this process, it might cover a set of structural anchors $\{c_1, c_5, c_8\}$. Since these are new, the check at line 14 is also true for each, granting additional rewards $r^{str}$ and adding them to the global $C_{cov}$ map. The agent continues this process, learning a "happy path" to complete the entire quest.

Across many subsequent episodes, this learned "path" remains a reliable source of semantic rewards. However, the structural rewards along this path quickly diminish to zero, as $\{c_1, c_5, c_8\}$ and other anchors on this path are now in $C_{cov}$. To continue maximizing its total reward, the agent is intrinsically motivated to deviate from its routine. It might try chopping the tomato on a different cutting board in a different place, or interacting with another kitchen tool before chopping, to see if these actions lead to new outcomes. Such exploratory actions, while potentially delaying the semantic reward, are incentivized by the prospect of discovering an uncovered anchor (e.g., $c_{12}$, an edge case in the chopping logic) and earning the $r^{str}$ bonus.

One concern of the adaptive hybrid reward design is the learning instability due to its \textit{unstable reward}. Referring to the studies of Curiosity-driven RL \cite{10.5555/3305890.3305968}, we address this challenge through several mechanisms. First, the semantic reward acts as a stable "backbone" for the learning process. It is consistently available in every episode, providing a clear and stationary objective that guides the agent towards a baseline level of functional competence. The structural reward, in contrast, serves as a diminishing bonus that catalyzes exploration without destabilizing the core policy. Second, we carefully balance the magnitudes of two types of rewards. By setting the one-time structural discovery bonus to be significant enough to incentivize deviation but not so large as to overshadow the cumulative semantic reward of completing the entire task, we manage the trade-off between exploration and exploitation. Finally, the underlying PPO algorithm itself incorporates mechanisms like entropy regularization, which intrinsically encourages policy stochasticity and prevents premature convergence, naturally complementing our external drive for exploration.

The adaptive structural reward calculation is realized through automated code instrumentation and tracking. We leverage standard dynamic analysis tools \texttt{pytest} to instrument the game's source code before the training process begins. This instrumentation adds lightweight probes that record the execution of specific lines and control-flow branches without altering the core game logic. During runtime, when the agent performs an action and the game engine executes the corresponding code, these probes are triggered. The set of activated probes within a single step is collected and passed back to the RL environment through the `info' dictionary. This allows the runtime engine to identify the set of newly covered anchors ($info[\text{'covered\_anchors'}]$) in real time and compute the structural reward accordingly. This entire process is fully automated, requires no manual code modification, and introduces minimal calculation overhead.

\section{Evaluation}
\label{sec:evaluation}

The evaluation aims to answer the following two research questions:
\begin{itemize}
    \item \textbf{RQ1: Comprehensiveness and Effectiveness.} How does the proposed \methodname{} framework compare in terms of testing comprehensiveness and effectiveness against a range of baseline methods?
    \item \textbf{RQ2: Efficiency and Cost.} How does the \methodname{} framework compare in terms of efficiency against baseline methods? 
    \item \textbf{RQ3: (Ablation Analysis)}: What are the respective contributions of the key components within the \methodname{} framework to its overall performance?
\end{itemize}

\subsection{Experiment Game Environment}

\begin{figure}[h!tb]
    \centering
    \includegraphics[width=1\linewidth]{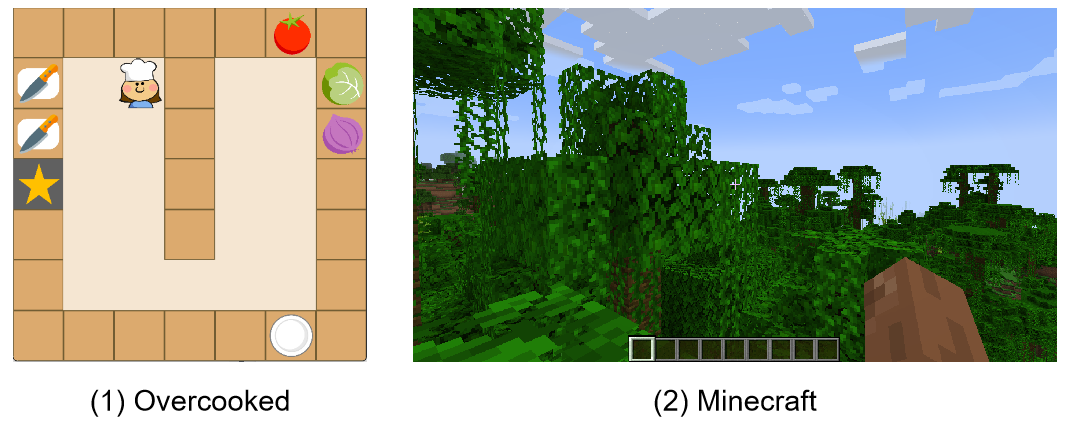}
    \caption{Snapshots of the experimental game environments.}
    \label{fig:gameimage}
\end{figure}
As shown in Figure~\ref{fig:gameimage}, we conduct experiments in two well-known games: Overcooked \cite{overcooked2016} and Minecraft \cite{minecraft2009} (Fig.~\ref{fig:gameimage}). Overcooked is a 2D simulation game centered on completing cooking tasks, where players act as chefs who interact with ingredients and kitchen tools to prepare various recipes. The game is widely celebrated for its cooperative design, earning multiple awards—including a BAFTA Games Award—and achieving significant commercial success with millions of copies sold. 
Given the closed-source nature of the original game, we employ its open-source experimental toolkit \cite{cai2024overcooked}.
Minecraft is an open-world 3D sandbox game that allows players to freely explore, build, and dismantle structures. Since its release, it has become one of the best-selling games of all time and has cultivated a massive global player community, consistently praised for its creative and open-ended gameplay.

\subsection{Experiment Settings}

\subsubsection{Comparison Baseline Methods}
\label{sec:baselines}

We compare \methodname{} against the following baselines.

\textit{Random Agent}.
This baseline represents a naive, uninformed exploration strategy, which is common in evaluating exploration strategies in complex environments~\cite{hu2024language}.. At each timestep, the agent selects an action uniformly at random from the entire set of available actions in the environment. It serves as a lower bound on performance, illustrating the results achievable without any form of intelligent guidance. 

\textit{PPO}. This baseline represents a standard, task-oriented RL approach. We use PPO~\cite{schulman2017proximal} , whose reward function is based solely on completing the quest. This agent is unaware of the underlying code changes and is only motivated to find an efficient policy to succeed at the task. It serves to demonstrate the performance of a pure "intent-only" testing paradigm. We utilize the robust implementation from the Stable-Baselines3 library~\cite{raffin2021stable}.

\textit{PPO+ICM (Curiosity-Driven)}.
This baseline augments the PPO with an Intrinsic Curiosity Module (ICM)~\cite{10.5555/3305890.3305968}. The ICM provides an additional intrinsic reward to the agent for visiting novel states, which are defined as states that the agent's internal forward dynamics model fails to predict accurately. This encourages the agent to explore unfamiliar parts of the state space, even if they do not directly contribute to the extrinsic task reward. This baseline represents a sophisticated, code-agnostic exploration strategy that aims to maximize state space coverage.

\textit{\methodname{} (Ours and its Ablations)}
This category includes our full proposed framework and several ablated versions designed to isolate the contributions of its key components, particularly the hybrid reward structure.

\begin{itemize}
    \item \textit{\methodname{} (Full)}: This is the complete implementation of our proposed framework as described in Section~\ref{sec:proposal}. The agent is guided by the adaptive hybrid reward function, which combines both sequential semantic rewards for subgoal completion and adaptive structural rewards for discovering new code coverage.

    \item \textit{\methodname{} (Semantic-Only)}: This ablation follows the same sequential curriculum of subgoals as the full method, but its reward function consists \textit{only} of the semantic rewards ($R_{sem}$) for completing each subgoal in order. The entire structural reward component ($R_{str}$) is removed. 

    \item \textit{\methodname{} (Structural-Only)}: In this ablation, the agent's reward function is based only on the adaptive structural reward ($R_{str}$) for covering previously unexecuted modified lines and branches. The semantic rewards for subgoal completion are completely removed. 

    \item \textit{\methodname{} (Global-Hybrid)}: This variant eliminates the hierarchical subgoal decomposition and the context-aware anchor mapping. Instead of a sequence of subgoals, the agent receives a semantic reward only upon the completion of the final quest (i.e., a sparse reward setting). Furthermore, the structural reward is globalized: the agent is incentivized to trigger \textit{any} previously uncovered anchor from the entire set $C$ at any timestamp, without restricting coverage to specific gameplay stages. 

\end{itemize}

\subsubsection{Settings in Overcooked}
The evaluation of Overcooked is based on an open-source implementation \cite{cai2024overcooked}, where the environment is a grid world with discrete actions (e.g., \textit{pick up tomato}, \textit{chop ingredient}, \textit{submit dish}), and tasks are constructed according to real Overcooked recipes. The state space includes the player’s position, held items, and the status of environment objects. We simulate real-world game update cases and design two core element updates (``dough” and ``oven”) from which 21 new tasks, approximately 200 lines of code differences, and around 100 branch differences are derived. (see Listing~\ref{lst:overcooked_tasks})

In the Overcooked environment, each task is trained for up to 100,000 environment interactions, with a fixed episode horizon of 300 steps. To encourage efficient behavior, the agent receives a per-step penalty of –0.1. To ensure a balanced reward scale, the terminal reward for completing the quest is fixed at 200, while intermediate semantic rewards for subgoals range between 10 and 50 depending on the subgoal complexity. 
To ensure the consistency of underlying reinforcement learning algorithms across different methods, we adopt the reinforcement learning suite provided by the Python \textit{stable-baselines3} library. We use the default MlpPolicy as the policy model, set the learning rate to 3e-4, collect 2048 timesteps per rollout, and use a discount factor \(\gamma = 0.99\).

For \textit{PPO+ICM} method, we incorporate an intrinsic-motivation signal based on state-visitation counts. At each step, the agent receives a curiosity bonus computed as
\(
r_{\text{int}}(s) = \frac{0.01}{\sqrt{N(s)}},
\)
where \(N(s)\) denotes the number of times state \(s\) has been visited. This intrinsic reward is added to the sparse extrinsic reward and jointly optimized via PPO to encourage exploration of novel states.


\subsubsection{Settings in Minecraft}
The Minecraft environment is used to evaluate the scalability of the proposed method in a more complex open-world setting. The player starts from a fixed position and can execute parameterized high-level commands such as “mine,” “craft,” and “attack,” each of which is automatically decomposed into a sequence of atomic operations. The environment state includes the player’s position, inventory, and the dynamic states of spatial elements (e.g., resource blocks, furnaces, and enemies). We simulate real-world game update scenarios and design one core element update (``gold update''), from which 20 new tasks, approximately 180 lines of code differences, and around 100 branch differences are derived. (see Listing~\ref{lst:mc_tasks})

For the Minecraft environment, we adopt the same reward shaping scheme but employ a larger training budget due to its more complex state and action spaces. Each task is allowed up to 1,000,000 environment interactions, with an episode horizon of 500 steps, while maintaining the same –0.1 per-step penalty. We've set the same reward levels as Overcooked, maintaining a base reward of 200 for completing the task, with interim rewards fluctuating between 10 and 50.
Additionally, we employ the same Python \textit{stable-baselines3} library for reinforcement learning and keep the policy model and hyperparameters identical to those used in Overcooked.

\subsection{Experiment Results and Discussion}

\begin{table*}[!t]
\centering
\caption{Comparative performance metrics on the Overcooked environment.}
\label{fig:overcooked_matrics_table}
\resizebox{\textwidth}{!}{%
\begin{tabular}{llllllll}
\toprule
Method &
\shortstack{Line\\Coverage $\uparrow$} &
\shortstack{Branch\\Coverage $\uparrow$} &
\shortstack{Unique States $\uparrow$} &
\shortstack{Ngram-3\\Diversity $\uparrow$} &
\shortstack{Anchor\\Discovery $\uparrow$} &
\shortstack{Success\\Rate $\uparrow$} &
\shortstack{Mean\\Length $\downarrow$} \\
\midrule
Random          & 0.31$\pm$0.01 & 0.35$\pm$0.01 & 13,862.13$\pm$238.73      & \textbf{63.39$\pm$2.64}  & 98.73$\pm$1.75  & 0.06$\pm$0.00 & 286.09$\pm$6.69 \\
PPO             & 0.55$\pm$0.01 & 0.58$\pm$0.01 & 23,171.25$\pm$311.17      & 18.53$\pm$0.19           & 165.43$\pm$5.92 & 0.65$\pm$0.02 & 45.49$\pm$1.06 \\
PPO+ICM         & 0.53$\pm$0.00 & 0.55$\pm$0.01 & \textbf{27,130.44$\pm$1,306.04} & 37.02$\pm$1.31   & 166.86$\pm$3.87 & 0.62$\pm$0.00 & 43.44$\pm$0.18 \\
\methodname -Full       & \textbf{0.93$\pm$0.02} & \textbf{0.98$\pm$0.01} & 23,730.02$\pm$676.49      & 28.86$\pm$0.88           & \textbf{289.71$\pm$1.40} & \textbf{0.98$\pm$0.01} & 54.92$\pm$2.26 \\
\methodname -Semantic   & 0.52$\pm$0.02 & 0.61$\pm$0.02 & 23,548.03$\pm$14.10       & 20.85$\pm$1.03           & 164.16$\pm$8.02 & 0.96$\pm$0.02 & \textbf{39.68$\pm$0.47} \\
\methodname -Structural & 0.31$\pm$0.01 & 0.28$\pm$0.01 & 13,754.21$\pm$485.12      & 10.87$\pm$0.19           & 90.19$\pm$3.08  & 0.05$\pm$0.00 & 286.50$\pm$3.18 \\
\methodname -Global-Hybrid & 0.29$\pm$0.01 & 0.30$\pm$0.03 & 14,486.92$\pm$712.85     & 12.98$\pm$0.13           & 95.37$\pm$3.78  & 0.05$\pm$0.01 & 282.41$\pm$2.99 \\
\bottomrule
\end{tabular}}
\end{table*}

\begin{table*}[!t]
\centering
\caption{Comparative performance metrics on the Minecraft environment.}
\label{fig:mc_matrics_table}
\resizebox{\textwidth}{!}{%
\begin{tabular}{llllllll}
\toprule
Method &
\shortstack{Line\\Coverage $\uparrow$} &
\shortstack{Branch\\Coverage $\uparrow$} &
\shortstack{Unique States $\uparrow$} &
\shortstack{Ngram-3\\Diversity $\uparrow$} &
\shortstack{Anchor\\Discovery $\uparrow$} &
\shortstack{Success\\Rate $\uparrow$} &
\shortstack{Mean\\Length $\downarrow$} \\
\midrule
Random
& 0.24$\pm$0.01 & 0.19$\pm$0.00
& 152{,}104.31 $\pm$ 90.43
& \textbf{98.75$\pm$3.09}
& 64.12$\pm$3.16
& 0.00$\pm$0.00
& 501$\pm$0.00 \\

PPO
& 0.46$\pm$0.01 & 0.51$\pm$0.00
& 224{,}947.61 $\pm$ 8{,}825.44
& 32.33$\pm$1.33
& 145.29$\pm$7.02
& 0.70$\pm$0.03
& 48.26$\pm$0.83 \\

PPO+ICM
& 0.50$\pm$0.01 & 0.51$\pm$0.02
& \textbf{243{,}554.39 $\pm$ 7{,}534.75}
& 35.83$\pm$1.42
& 149.43$\pm$0.22
& 0.69$\pm$0.03
& 54.81$\pm$2.29 \\

\methodname -Full
& \textbf{0.92$\pm$0.01} & \textbf{0.94$\pm$0.01}
& 210{,}947.41 $\pm$ 3{,}804.29
& 22.23$\pm$0.18
& \textbf{273.78$\pm$3.32}
& \textbf{0.98$\pm$0.01}
& 69.69$\pm$1.65 \\

\methodname -Semantic
& 0.54$\pm$0.03 & 0.52$\pm$0.03
& 233{,}949.12 $\pm$ 7{,}850.75
& 21.23$\pm$0.34
& 159.19$\pm$6.48
& 0.97$\pm$0.01
& \textbf{47.68$\pm$2.09} \\

\methodname -Structural
& 0.24$\pm$0.00 & 0.16$\pm$0.01
& 96{,}842.19 $\pm$ 186.63
& 7.15$\pm$0.08
& 67.79$\pm$2.41
& 0.00$\pm$0.00
& 501$\pm$0.00 \\

\methodname -Global-Hybrid
& 0.25$\pm$0.01 & 0.18$\pm$0.03
& 118{,}163.18 $\pm$ 3{,}899.65
& 14.74$\pm$0.55
& 59.31$\pm$0.71
& 0.00$\pm$0.00
& 501$\pm$0.00 \\

\bottomrule
\end{tabular}}
\end{table*}

The experimental results (see Table~\ref{fig:overcooked_matrics_table} and Table~\ref{fig:mc_matrics_table}) show the overall advantages of our proposal while revealing the fundamental limitations of traditional reinforcement learning in testing scenarios. Although PPO and PPO\_ICM achieve a reasonable level of task success rate (approximately 0.7), their line coverage remains at only 0.46--0.55, and their mean anchor discovery is extremely low. This indicates that task-oriented agents tend to complete tasks via the ``shortest path,'' thereby systematically ignoring the non-critical paths (edge cases) introduced during updates. In contrast, \methodname{}-Full, utilizing a hybrid reward mechanism, increases both line and branch coverage to 0.9/0.95.  This strongly demonstrates the core idea of our proposal that only by explicitly incorporating code structure changes into the reward function can the agent be truly guided into the ``code space'' under test.

\begin{figure}[h!tb]
    \centering
    \includegraphics[width=\linewidth]{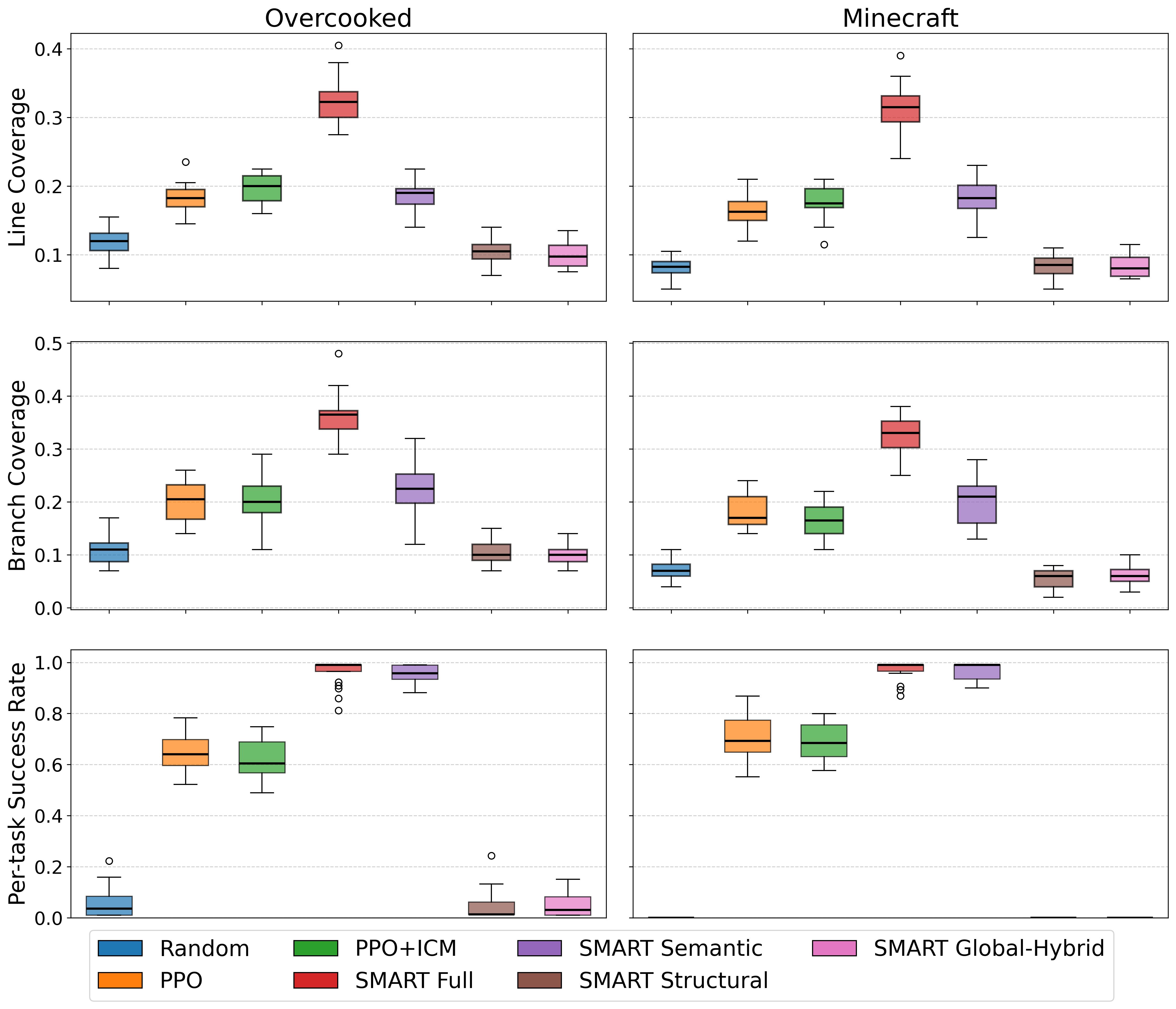}
    \caption{Detailed distribution of structural coverage and task success rates.}
    \label{fig:metric_1}
\end{figure}

Regarding the performance of the curiosity mechanism (ICM), although PPO\_ICM achieves a slightly higher number of unique states (27.1k) than \methodname{} (as shown in Table~\ref{fig:overcooked_matrics_table}), its discovery rate for new code (166.86) is much lower than that of \methodname{}. This suggests a mismatch between ``state novelty'' and ``code novelty'': ICM encourages the agent to explore visually distinct or dynamically unpredictable states (such as running around the map or viewing different scenery), but these states do not necessarily correspond to the modified code branches in the current code update (about 200 lines of diff). Therefore, blindly maximizing state diversity is an inefficient strategy for testing specific code updates. In contrast, \methodname{}'s exploration is ``constrained and precise'': although the total number of states it visits is slightly lower, every new state explored closely revolves around code change points (anchors). This demonstrates that, in incremental testing scenarios, code-aware directed exploration (coverage-aware exploration) is far more cost-effective than generic exploration based on pixel or state prediction error (curiosity-driven exploration). Additionally, with regard to Ngram-3 diversity, Random is the strongest, followed by ICM, indicating that mere diversity in action sequences does not directly translate into effective testing of updated code.
Ablation studies further analyze the contributions of each component. First, as shown by the distribution in Figure~\ref{fig:metric_1}, \methodname{} - Semantic achieves a high success rate (about 97\%) and the shortest average trajectory length, but its code coverage is significantly lower than that of the full method (for example, 52\% line coverage vs. 93\% in Overcooked). This result actually highlights the ``happy path'' problem: while LLM-generated semantic rewards can perfectly guide the agent to complete the task, they often lead it down the most standard and efficient path, effectively bypassing error-handling logic, boundary checks, and other defensive programming regions.

In contrast, the complete failure of \methodname{} - Structural (with both success rate and coverage near random levels, and frequent timeouts, see Table~\ref{fig:mc_matrics_table}) confirms that the ``reachability'' of deep code logic heavily depends on high-level game context. Without semantic subgoal guidance from the LLM, the agent cannot construct long sequences of dependent actions, resulting in code logic deeply embedded at the backend of the task being physically unreachable.

Equally noteworthy is the poor performance of \methodname{} - Global-Hybrid. Although this variant retains both semantic goals and structural reward signals, its results (low success rate and coverage comparable to random) further validate the necessity of goal decomposition and context mapping. Without decomposing complex tasks into a sequence of subgoals, agents struggle to overcome the challenge of sparse rewards; and lacking context mapping to structural anchors, global code coverage rewards become too sparse and lack directionality. This indicates that a simple additive combination of ``functional + structural'' rewards is ineffective. Only the fine-grained alignment proposed by \methodname{} enables structural signals to truly assist functional exploration.

\begin{figure}[h!tb]
    \centering
    \includegraphics[width=1\linewidth]{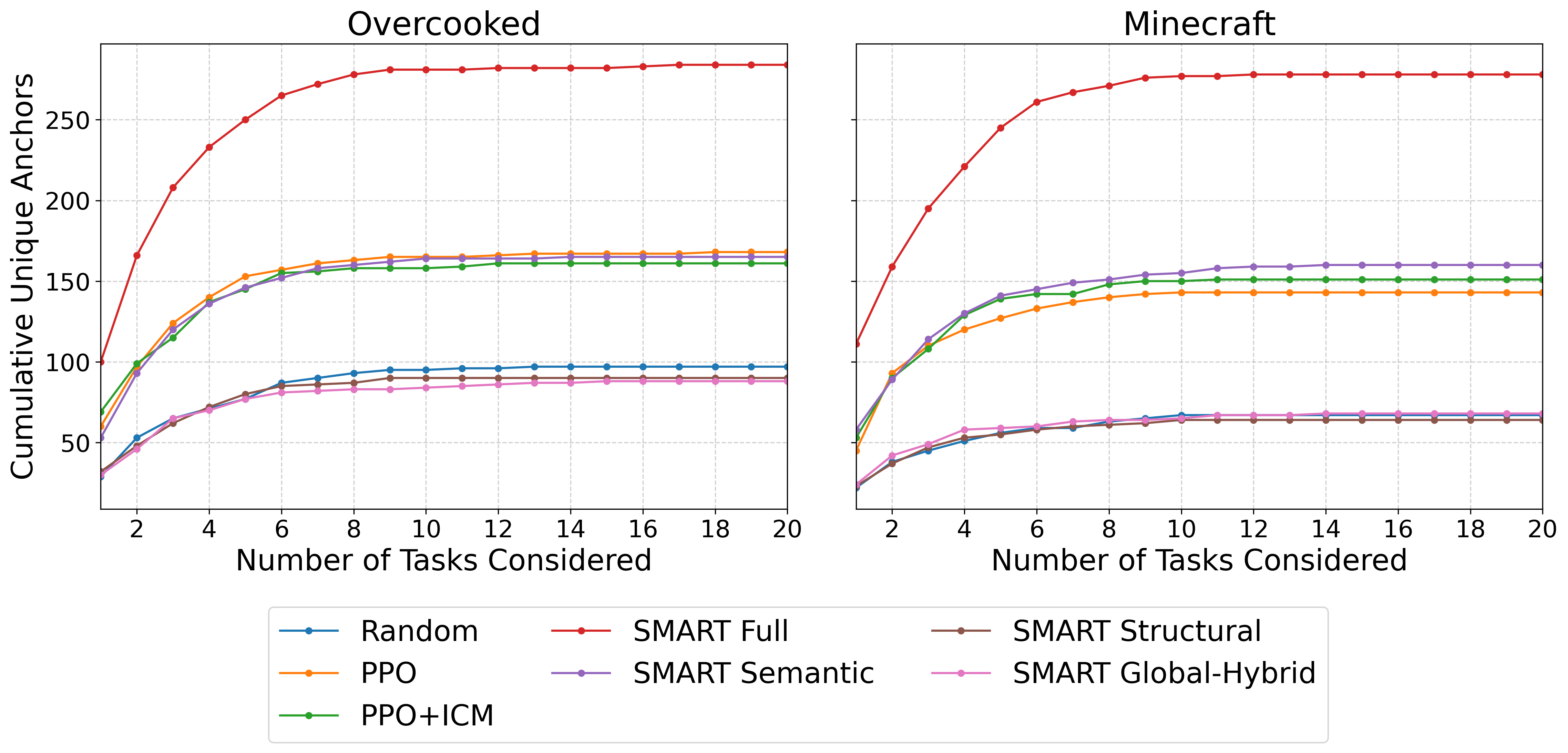}
    \caption{Learning curves for cumulative structural anchor coverage.}
    \label{fig:metric_2}
\end{figure}

Finally, the full method (\methodname{} - Full) results in slightly longer trajectories than the semantic-only variant (e.g., 69.69 vs. 47.68 in Minecraft), but with significantly higher coverage. This increase in trajectory length represents ``productive inefficiency'': the structural reward mechanism successfully incentivizes the agent to deviate from the optimal path and perform exploratory actions (such as trying different tools or materials), thereby triggering edge-case code. In traditional RL tasks, an increase in steps is usually regarded as a performance decline, but in testing scenarios, this deliberately extended trajectory essentially reflects the agent trading time cost for higher anchor discovery (as shown in Figure~\ref{fig:metric_2}), which is precisely the behavior pattern desired in automated testing to balance functional and structural verification.

Figure~\ref{fig:metric_1} further presents box plots of structural coverage and success rates under different task granularities. A detailed analysis of the statistical features in this figure yields two key conclusions. First, the performance advantage of \methodname{} demonstrates a high degree of universality, i.e., \methodname{} achieves significantly higher code line and branch coverage across the vast majority of individual tasks. Second, \methodname{} exhibits robustness, i.e., observing the interquartile range of the box plots, \methodname{} shows a compact coverage distribution and a high lower quartile, indicating the absence of obvious ``bottleneck'' tasks in the test set. Even when faced with tasks involving complex logic or intricate interaction steps---which are typically blind spots for methods like PPO---\methodname{} is able to maintain a high level of structural coverage. 

Figure~\ref{fig:metric_2} depicts the growth trend of the cumulative number of unique code anchors covered as the number of test tasks increases (tasks are sorted by ascending complexity). It is evident that \methodname{} features the steepest growth curve, indicating that for each new test task introduced, \methodname{} can efficiently discover a large number of previously untriggered code logic branches. In contrast, baseline methods encounter an early ``plateau,'' suggesting a tendency to reuse existing simple interaction patterns (happy paths) when facing new tasks, which makes it difficult to reach deep code branches unique to the new tasks. Furthermore, as task complexity increases (rightward along the X-axis), the gap between \methodname{} and the baselines widens, reaffirming that \methodname{} holds significant scalability advantages in incremental development settings.

Although \methodname{} achieves significantly higher coverage than baseline methods in both game environments, its final coverage still falls short of the theoretical 100\%, as illustrated in Figure~\ref{fig:metric_2}. We conducted an in-depth code-level review of these uncovered anchors and identified two main contributing factors. 
First, some of the structural anchors identified by AST differencing correspond to extreme defensive programming or exception-handling logic. For example, in the ``Onion Pizza'' task of Overcooked, certain uncovered code sections pertain to protective checks for ``illegal ingredient stacking order'' or ``forced interaction after prolonged idle baking.'' Guided by the normal task semantics and LLM-generated subgoals, the agent's behavior remains within logical bounds, and thus rarely triggers these fallback mechanisms designed to prevent program crashes.
Second, this outcome represents a necessary trade-off between training efficiency and exploration depth. Under constraints of limited training steps and per-step penalties, PPO-based strategies naturally converge to paths with more stable reward returns. Although our structural reward ($R_{str}$) provides strong incentives for exploratory behaviors, systematically covering all edge-case branches (e.g., deliberately waiting until a timeout to trigger an error prompt) may lead the agent to frequently deviate from the task objective, significantly reducing training efficiency and destabilizing policy convergence. Thus, \methodname{} is designed to maximize structural coverage while ensuring thorough testing of the main functional paths and common interaction logic, rather than pursuing exhaustive coverage of all dead code or unreachable logic at any cost.
To address these residual blind spots, a promising solution is to integrate \methodname{} with fuzz testing or symbolic execution techniques. While \methodname{} excels at verifying complex, long-dependency gameplay logic, these complementary techniques can ``brute-force'' trigger counter-intuitive defensive branches and exception-handling code via constraint solving or random mutation. Together, they can construct a comprehensive automated testing framework that covers both happy paths and edge cases.

\subsection{Limitations}
Although our proposed \methodname{} framework demonstrates promising results in synergizing structural code with functional gameplay validation, several methodological limitations warrant discussion.

First, the effectiveness of our framework is highly dependent on the quality and accuracy of the semantic intent generated by the LLM. The core innovation of \methodname{} lies in its ability to translate low-level code modifications into high-level, testable gameplay objectives. However, If the LLM misinterprets the purpose of a code change---for example, by hallucinating a gameplay feature from a simple refactoring, or failing to capture the subtle nuances of a balance adjustment---the resulting semantic reward function may be misaligned. A flawed semantic reward could potentially drive the RL agent toward irrelevant or incorrect behaviors, thus undermining the very synergy between structural and functional testing that \methodname{} aims to achieve.

Second, similar to existing agent-based testing methods, this paper focuses on testing updates to the core gameplay logic of the game. Therefore, we employ AST analysis, which excels at identifying modifications in control flow, data structures, and algorithmic computations. However, we acknowledge that modern games are increasingly complex and data-driven. For example, changes to user interface (UI) layouts defined in XML files, adjustments to 3D model assets, modifications in shader code for graphical effects, or tweaks in neural network-based AI behavior often do not manifest as interpretable changes in the primary source code's AST. Updates outside the core game logic thus require complementary techniques from other domains to be effectively tested.

Third, it is important to emphasize that \methodname{} is designed for frequent, incremental testing rather than as a comprehensive replacement for the thorough validation required in major releases. As previous work notes, robust quality assurance---especially for large updates---relies on a combination of regression testing, performance profiling, and manual exploratory testing to ensure overall game stability~\cite{Politowski2021A}. While \methodname{} focuses on change deltas and does not achieve exhaustive system-wide verification, it provides unique value when integrated into a broader testing pipeline. Specifically, \methodname{} offers two key contributions: First, it acts as an intelligent, automated smoke test, rapidly validating the most critical functionalities affected by updates. This gives QA teams an early signal of build stability before committing to full regression cycles. Second, the test trajectories generated by the \methodname{} agent serve as high-quality, machine-generated assets that already validate new features. These trajectories can be curated and optimized to seed new regression test cases, accelerating the maintenance of the test suite and reducing the manual effort required to create new validation scenarios from scratch.

Fourth, we specifically focus on the element addition scenario in the experiment, as the introduction of new gameplay elements and associated quests constitutes the most frequent and representative update pattern in the GaaS model. Although real game version updates typically involve substantial and diverse content changes, this decision is grounded in operational reality and based on two considerations:
First, existing automated game testing baselines, particularly those agent-based methods, are generally designed and evaluated for a specific, singular game task or quest. By restricting the experiment to tasks triggered by element update, we can eliminate interference from multi-task scheduling, thereby allowing for a more precise and fair comparison with baseline methods in terms of structural coverage and functional completion on the same dimension.
Second, rather than focusing on external version updates, our method assumes an internal Continuous Integration (CI) process. In this context, "Gameplay Intent" is not monolithic but composed of discrete functional goals. A single code commit typically involves a small number of changes (such as adding a new element and its corresponding quests), representing an atomic unit of gameplay intent. Since a large studio may execute dozens of such commits daily, testing a single element update corresponds to verifying the most fundamental and frequent unit of intent. Therefore, ensuring the completion of the specific task associated with the code update is effectively validating the "Gameplay Intent" inherent in that specific increment.
To address complex externally released updates that combine multiple elements, a natural extension strategy is "hierarchical decomposition and code anchoring," which involves decomposing a large update into multiple such atomic units.

Fifth, the design of the hybrid reward function introduces a significant challenge in balancing the structural and semantic reward components. Specifically, the relative weighting of these two signals is non-trivial and may be highly context-dependent. An imbalance could lead the agent to "game" the system; for instance, it might prioritize executing many shallow code paths to maximize the structural reward, while failing to adequately validate the complex, multi-step functional intent of an update. This may require careful, per-update tuning of reward weights, which could introduce a layer of manual oversight and potentially compromise the framework's full automation. Future work could explore more advanced reward-shaping mechanisms to address this challenge dynamically.

Finally, our reliance on AST-based static analysis inherently limits the detection of dynamic code behaviors. Modern game engines often utilize reflection, dynamic asset loading, or hot-swapped scripts (e.g., Lua or Python) whose control flow cannot be fully resolved statically. If a game update relies heavily on such dynamic dispatch mechanisms, the structural anchor mapping in Stage 4 may fail to identify or link these code paths to the correct subgoals. Therefore, it is also a promising direction to integrate dynamic call-graph generation to complement the static AST differencing.


\section{Related Work}
\label{sec:relatedwork}
This section reviews two core research directions in the field of automated game testing, and discusses the position of this study.

\subsection{Structural Game Testing}

With the rise of the GaaS, games require frequent content updates and feature iterations, making regression testing an indispensable part of software quality assurance~\cite{RLReg}. The goal of regression testing is to verify that new code changes do not inadvertently break existing functionality.

Early automation efforts mainly focused on record-and-replay and script-based testing. The record-and-replay mechanism captures human testers' action sequences and transforms them into reusable test cases, reducing repetitive labor~\cite{ostrowski2013automated, mioto2025mapping}. Similarly, script-based testing relies on testers writing control scripts to simulate player behavior, which is suitable for verifying fixed game scenarios and task flows~\cite{spronck2006adaptive, cho2010scenario}. However, both approaches have significant limitations. They are heavily reliant on manual creation and maintenance of test assets; when game UIs or functional logic change, these scripts or recorded traces often become invalid, resulting in high maintenance costs~\cite{ixie2024gametesting}.

To improve regression testing efficiency, researchers have introduced more advanced Regression Test Selection (RTS) techniques, whose core idea is to execute only the test cases potentially affected by code changes~\cite{rothermel1996safe}. For instance, GameRTS~\cite{GameRTS} constructs a game state transition graph through static code analysis, and after version updates, it intelligently identifies the most relevant test cases for re-testing by analyzing code differences. This approach effectively avoids unnecessary test executions while maintaining a high defect detection rate.

However, these code-centric regression testing methods have a fundamental limitation: they focus on structural verification but lack an understanding of functional intent. For example, in a cooking game like \textit{Overcooked}, an update might modify the interact function to support a new ``chopping'' mechanic for tomatoes. An RTS tool can accurately identify this code change and select a generic test case where the avatar simply triggers the interaction button. Nevertheless, it cannot determine the underlying intent—whether the interaction is meant to transform a whole tomato into a chopped state as part of a new ``Pizza'' recipe, or simply to pick up the item. Consequently, it cannot select test scenarios specifically aimed at verifying if the chopping action actually progresses the specific culinary task or if the state transition logic functions correctly within the recipe workflow. These methods can only confirm that the code structure has been exercised, but cannot verify whether the intended higher-level gameplay goals of the structural changes have been achieved.

\subsection{Intent-Driven Game Testing}
To overcome the rigidity and high maintenance costs of traditional script-based methods, the research community has gradually shifted towards using agents capable of autonomously interacting with the environment for game testing. This approach emphasizes exploratory testing, aiming to discover unintended behaviors and edge cases.

Early agent-based methods include model-based testing, where a formal model of gameplay is constructed in advance, and agents generate test paths according to the model~\cite{iftikhar2015automated, becares2017approach}. Additionally, behavior trees have been used to build agents for level evaluation and playability analysis~\cite{stahlke2020artificial, stahlke2019artificial}. These methods offer more flexibility than fixed scripts but still require substantial upfront design and modeling effort.

The emergence of RL has brought a paradigm shift to automated game testing. RL agents learn gameplay through trial-and-error, without the need for predefined scripts or models, making them highly suitable for exploring vast and dynamic game worlds~\cite{bergdahl2020augmenting}. To enhance the efficiency and breadth of exploration, researchers have proposed various augmentations. Among these, curiosity-driven approaches introduce intrinsic rewards to incentivize agents to visit novel or unexplored game states, thereby significantly improving state-space coverage and the ability to uncover hidden defects~\cite{9619048}. Moreover, some studies combine deep reinforcement learning with evolutionary strategies, evolving diverse agent policies to balance task completion and state exploration, which has shown promising results in complex online battle games, such as the Wuji framework~\cite{zheng2019wuji}. To make agent behavior more closely resemble real players, persona-based testing has been developed. By assigning agents different predefined personas (e.g., ``explorer'' or ``achiever'') and designing corresponding reward functions, agents are guided to execute test paths aligned with specific motivations~\cite{holmgaard2018automated, ariyurek2021playtesting}.

Although these agent-based methods excel at simulating player behavior and functional validation, they typically operate at the intent level and lack direct connections to underlying structural changes. Most run in black-box or gray-box environments, with behavior driven by high-level game feedback (e.g., scores, task completion status). A fundamental limitation of this reward-driven approach is the agent's tendency to converge on a single optimal strategy, often referred to as the Happy Path.'' For instance, an RL agent trained to complete a cooking quest will learn the most efficient sequence of actions to deliver a dish. Consequently, it will systematically avoid suboptimal interactions or erroneous states—such as burning an ingredient or combining invalid items—effectively bypassing the very defensive programming and exception-handling logic introduced in the code update. As a result, while the agent proves the task is clearable, it fails to exercise the alternative branches and edge cases that constitute a significant portion of the structural changes. 

\subsection{Position of This Paper}
The preceding review highlights a key dichotomy in current game testing methodologies: the gap between low-level structural verification and high-level functional/intent validation. Traditional regression testing approaches~\cite{GameRTS, ostrowski2013automated} excel at analyzing changes from the code perspective and conducting structural testing, but they are unable to comprehend the higher-level functional intent behind code modifications, nor can they proactively verify whether these intents have been correctly implemented. In contrast, agent-based exploratory or behavior-driven testing~\cite{9619048, zheng2019wuji, ariyurek2019automated} focuses on simulating player behavior and validating functionality. However, due to the lack of direct association with underlying code changes, these methods cannot guarantee effective coverage of specific code modifications, and thus may miss regression defects that affect only code structure without immediately altering macro-level behavior.

The \methodname{} framework proposed in this study aims to systematically bridge this critical gap. Unlike previous work, \methodname{} does not merely utilize LLMs for general game task planning~\cite{hu2024language, enhong_IEICE25} or test case generation~\cite{paduraru2024unit, taesiri2022large,xu2024largelanguagemodelssynergize}. Instead, the core novelty \methodname{} lies in directly grounding itself in the most fundamental structural changes—source code differences (AST diff)---and leveraging the powerful semantic understanding capabilities of LLMs to automatically reverse-engineer and interpret the high-level functional intent behind these structural modifications. Building upon this, \methodname{} introduces a novel hybrid reward function that integrates both the structural coverage reward derived from code changes and the semantic intent reward inferred by LLMs. This mechanism ensures that reinforcement learning agents are not only incentivized to execute all modified code paths but are also guided to align their behavior with the higher-level functional goals implied by these changes. 

\section{Conclusion and Future Work}
\label{sec:conclusion}

In this paper, we proposed \methodname{}, a novel automated testing framework that synergizes code coverage with gameplay intent. By leveraging LLMs to interpret AST differences and decomposing them into semantic subgoals, \methodname{} constructs a context-aware hybrid reward system. This system guides RL agents to not only fulfill the functional requirements of new game updates but also to actively explore and cover the underlying structural modifications. Our experiment evaluation on \textit{Overcooked} and \textit{Minecraft} demonstrates that \methodname{} significantly outperforms traditional RL and curiosity-driven baselines, achieving an extremely high coverage of modified code branches while maintaining high task success rates.

In future work, we plan to expand \methodname{} in three directions. First, we aim to implement a closed-loop refinement mechanism to mitigate the impact of LLM hallucinations in one-off subgoals and rewards. Specifically, we plan to introduce an iterative cycle where the agent's failure trajectories or low-coverage reports are fed back to the LLM, allowing it to dynamically debug and refine its generated semantic rewards and anchor mappings, thereby improving robustness against ambiguous code changes.
Second, we aim to address the challenge of non-code updates by incorporating multi-modal analysis, allowing the framework to test changes in game assets, UI layouts, and data tables that do not manifest in the AST. 
Finally, acknowledging the trade-off between RL exploration and deep defensive logic coverage, we will explore hybridizing \methodname{} with symbolic execution or fuzz testing.


\lstset{
  backgroundcolor=\color{gray!5},   
  frame=single,                     
  rulecolor=\color{gray!60},        
  breaklines=true,                  
  breakatwhitespace=true,           
  columns=flexible,                 
  keepspaces=true,                  
  aboveskip=0.5em,                    
  belowskip=0.5em,                    
  captionpos=b,                     
  basicstyle=\ttfamily\scriptsize, 
}

\section*{Appendix}
This appendix provides detailed prompts, output examples from the \methodname framework stages, and the exmaple lists of tasks used in the experiments.

\begin{lstlisting}[
    caption={Prompt for Subgoal Generation (Stage 2).},
    label={lst:prompt2}
]
{
  "system_role": "You are a subgoal generator and structural anchor annotator.",
  "task_description": {
    "summary": "Given an AST difference report (ast_diff.txt) generated by a static analysis tool, identify newly added or significantly expanded game tasks, decompose them into ordered subgoals, and annotate each subgoal with structural anchors.",
    "steps": [
      "Identify newly added or significantly expanded game tasks.",
      "Decompose each complex task into an ordered sequence of subgoals S = (sg1, ..., sgN), where each subgoal is a natural-language description, verifiable, and relatively atomic.",
      "Annotate each subgoal with structural anchors-specific code locations directly related to that subgoal."
    ]},
  "input_data": {
    "source": "AST difference report file (ast_diff.txt) generated by a static analysis tool.",
    "content_format": "Text diff containing module headers and unified diff lines.",
    "ast_diff_wrapper": {
      "begin_marker": "===== BEGIN AST DIFF =====",
      "end_marker": "===== END AST DIFF =====",
      "placeholder": "{ast_diff_text}"},
    "module_header_example": "== Module: controller/action.py ==",
    "line_type_description": {
      "added_lines": "Lines beginning with '+' (new version). Prefer these as anchors.",
      "removed_lines": "Lines beginning with '-' (old version). Use only when describing removed behavior."}},
  "task_definitions": {
    "new_or_expanded_task": "A new task or a significantly expanded existing task inferred from the AST diff. Tasks generally correspond to noticeable flows, gameplay steps, crafting recipes, combat logic, or prerequisite logic.",
    "subgoal": {
      "description": "A subgoal is an ordered step within a task, described in natural language, with clear verifiable conditions and relatively atomic behavior.",
      "requirements": [
        "Natural-language description.",
        "Verifiable in code or gameplay logic.",
        "Relatively atomic (should not require further splitting into independent steps)."]},
    "structural_anchor": {
      "description": "A structural anchor is a set of code locations most directly related to a subgoal.",
      "format": "(file_path, line_number_array) pairs, where file_path comes from module headers and line_number_array uses unified diff line numbers."}},
  "output_requirements": {
    "overall": "For each new or expanded task, produce an object containing ordered subgoals and their structural anchors.",
    "task_object": {
      "fields": [
        "task_id: A unique string identifying the task (derived from code context or inferred task name).",
        "subgoals: An ordered list of subgoal descriptions.",
        "anchors: An array aligning with subgoals by index, each containing code anchor locations."]},
    "subgoal_properties": [
      "Ordered: sg1 -> sg2 -> ... -> sgN follow the completion sequence.",
      "Logical dependency: Later subgoals typically require earlier ones.",
      "Natural-language text."
    ],
    "structural_anchor_rules": {
      "anchor_format": "(file_path, line_number_array)",
      "file_path": "From diff module headers (e.g., 'controller/action.py').",
      "line_numbers": "From unified diff lines starting with '+' or '-'; prefer '+'.",
      "span_rules": [
        "Use [start, end] for multi-line spans, e.g., [424, 431].",
        "Use a single-element array for single lines, e.g., [188]."],
      "selection_preference": [
        "Prefer core logic: material requirements, action logic, crafting tables, attack logic, prerequisite checks, etc.",
        "Provide 2~5 anchors per subgoal; only leave empty when absolutely necessary."]},
    "json_output_format": {
    "description": "Output must be a single JSON object. Example:",
    "template": {
        "tasks": [{
        "task_id": "Craft_Iron_Sword",
        "subgoals": ["Subgoal 1", "Subgoal 2"],
        "anchors": [
            {"index": 0, "text": "Subgoal 1", "locations": [
                    {"file": "env/vector.py", "span": [424, 431]},
                    {"file": "controller/action.py", "span": [172, 182]}
                ]},
            {"index": 1, "text": "Subgoal 2", "locations": [
                    {"file": "env/vector.py", "span": [461, 476]}
                ]}]}]}}},
  "constraints": {
    "anchors_per_subgoal": {
      "min": 2, "max": 5,
      "note": "Leave empty only when anchor lines cannot be identified."},
    "line_usage_preference": [
      "Prefer '+' (new) version line numbers.",
      "Use '-' only when describing removed behavior."
    ],
    "mapping_precision": "Anchors should precisely correspond to the implementation logic of each subgoal.",
    "language": "All subgoals must be written in natural language.",
    "format_adherence": "Output must be strictly the specified JSON object without additional commentary."
  },
  "examples": {
    "intuitive_examples": [
      "Pizza: cut tomatoes -> make sauce -> assemble pizza -> bake pizza",
      "Craft_iron_Sword: collect wood -> obtain iron ore -> smelt iron -> craft iron sword"
    ],
    "note": "Actual subgoals must be strictly inferred from the AST diff ({ast_diff_text})."
  }
}

\end{lstlisting}

\begin{lstlisting}[
    caption={Prompt for Semantic Reward Generation (Stage 3).},
    label={lst:prompt3}
]
{
  "system_role": "You are a reward-function generation assistant, specialized in designing appropriate reward functions for game tasks.",
  "task_description": "You will receive a task, its subgoals, and various information extracted from the game code. Based on these inputs, you must generate multiple reward functions that guide the agent toward completing the task.",
  "input_data": {
    "description": "The input consists of a task ID, a list of subgoals, an item list, and a task list. These are injected into the placeholders {task_id}, {subgoals}, {ITEMS}, and {TASK_IDS}.",
    "example": {
      "task_id": "task_001",
      "subgoals": ["collect wood", "craft axe", "cut trees"],
      "ITEMS": ["wood", "stone", "axe"],
      "TASK_IDS": ["subtask_1", "subtask_2", "subtask_3"]}},
  "output_instructions": {
    "format": "You must generate a strict JSON object containing the target task ID and a list of reward definitions.",
    "json_template": {
      "target_task_id": "{task_id}",
      "rewards": [{"items": [["item_name", 1]], "reward": 10},
        {"task": "subtask_id", "reward": 15}]}},
  "critical_constraints": [
    {"constraint": "Full Process Coverage", "description": "Reward functions must cover the entire task process, from beginning to completion."},
    {"constraint": "Item-Based Reward Logic", "description": "For item-triggered rewards, the reward activates only when the inventory simultaneously contains all specified items in the required amounts."},
    {"constraint": "Subtask-Based Reward Logic", "description": "For subtask-triggered rewards, the reward activates upon completion of the specified subtask. You may include rewards for prerequisite subtasks to guide the agent."},
    {"constraint": "Once-Only Trigger", "description": "Each reward triggers only once. The agent incurs a 0.1 penalty per step."},
    {"constraint": "Granularity", "description": "Use fine-grained rewards to guide incremental progress. Staged rewards are allowed when collecting multiple quantities of items."},
    {"constraint": "Reward Scale", "description": "Main task reward is fixed at 200. Sub-rewards should range between 10 and 50."},
    {"constraint": "Format Adherence", "description": "Output must be strictly in JSON format, without any additional commentary."}
  ]
}

\end{lstlisting}

\begin{lstlisting}[
    caption={Prompt for Structural Anchor Mapping (Stage 4).},
    label={lst:prompt4}
]
{
  "system_role": "You are an anchor alignment assistant responsible for aligning subgoals in Minecraft tasks to specific code change locations (file path and line number ranges).",
  "task_description": "Use the provided code-reading tools to verify project structure and code changes, and map each task subgoal to precise code locations in the modified Python source files. All line numbers must come from actual tool outputs; no imaginary code or invented spans are allowed.",
  "tools_description": {
    "tree_project": "View the Python file tree of the project. Some irrelevant files such as pppo.py and env/vector.py may be excluded automatically.",
    "read_file": "Read the content of a Python source file by its relative path.",
    "search_project": "Search for a string across all Python files and return matches with surrounding context.",
    "list_diffs": "List the modules changed in this round of code modifications.",
    "file_diff": "Display a human-readable AST diff of a module, showing added, modified, or removed functions and fields, along with their old and new line numbers.",
    "change_spans": "Given a module path and a symbol name, return the old and new line-number spans (span_old / span_new)."
  },
  "output_instructions": {
    "format": "Produce a JSON object describing the code locations corresponding to each subgoal of the current task.",
    "json_template": {
      "task_id": "<current task ID>",
      "subgoals": [{
        "index": 0,
        "text": "original subgoal text (copied verbatim)",
        "locations": [{
          "file": "path/to/file.py",
          "span": [10, 11, 12],
          "kind": "function or non_function (optional)",
          "change_type": "added / modified / removed (optional)",
          "symbol": "function_or_variable_name (optional)",
          "qualname": "fully.qualified.name (optional)"
        }]}]}},
  "critical_constraints": [
    {"constraint": "Do Not Invent Code", "description": "All file paths and line number spans must come from tool results such as change_spans or file_diff."},
    {"constraint": "Accurate Span Usage", "description": "The span field must contain new-version line numbers. Prefer using line_numbers_new from change_spans whenever possible."},
    {"constraint": "Faithful Subgoal Mapping", "description": "Subgoals must be copied verbatim, and each mapped location must correspond to verified code changes relevant to that subgoal."},
    {"constraint": "Empty Allowed", "description": "If no code change corresponds to a subgoal, the locations list may be empty."},
    {"constraint": "JSON-Only Output", "description": "The final response must contain only the required JSON object, with no extra commentary."}
  ],
  "notes": [
    "Subgoal indices start from 0.",
    "Use the tools to fully understand the code before generating the final JSON.",
    "env/vector.py will not appear in list_diffs, even if changed.",
    "Prioritize change_spans for accurate symbol-level line ranges."]
}
\end{lstlisting}

\begin{lstlisting}[
    caption={Example Output of Subgoal Sequence (Stage 2).},
    label={lst:subgoal_example}
]
{"tasks": [{"task_id": "pizza_onion",
      "subgoals": ["Pick up a clean plate from a plate tile.",
        "Pick up a dough from a dough tile and place it on a counter so it can be used as a pizza base.",
        "Pick up a tomato from a tomato tile, bring it to a knife tile, and chop it until the chop progress is complete.",
        "Pick up cheese from a cheese tile and place both chopped tomato and cheese onto the dough on the counter so that the plate can later contain a pizza.",
        "Pick up the raw pizza on a plate and put it into an oven tile; wait until the cooked flag of the pizza becomes true without letting it burn.",
        "Take the cooked pizza out of the oven on a plate and, if the map provides onion, optionally chop onion on a knife and add it onto the same plate as an extra topping.",
        "Deliver the plate with the finished pizza (and any extra toppings such as onion) at a delivery tile so that TaskManager marks the task as correctly completed."],
      "anchors": [{"index": 0, "text": "Pick up a clean plate from a plate tile.",
          "locations": [{"file": "overcooked/overcookedPlus/module/item_manager.py", "span": [55, 118]},
            {"file": "overcooked/overcookedPlus/items.py", "span": [682, 780]},
            {"file": "overcooked/overcookedPlus/module/event_manager.py", "span": [180, 260]}]},
        {"index": 1, "text": "Pick up a dough from a dough tile and place it on a counter so it can be used as a pizza base.",
          "locations": [{"file": "overcooked/overcookedPlus/module/item_manager.py", "span": [60, 130]},
            {"file": "overcooked/overcookedPlus/items.py", "span": [260, 340]},
            {"file": "overcooked/overcookedPlus/module/event_manager.py", "span": [320, 480]}]},
        {"index": 2, "text": "Pick up a tomato from a tomato tile, bring it to a knife tile, and chop it until the chop progress is complete.",
          "locations": [{"file": "overcooked/overcookedPlus/module/item_manager.py", "span": [40, 90]},
            {"file": "overcooked/overcookedPlus/items.py", "span": [120, 220]},
            {"file": "overcooked/overcookedPlus/module/event_manager.py", "span": [520, 620]}]},
        {"index": 3, "text": "Pick up cheese from a cheese tile and place both chopped tomato and cheese onto the dough on the counter so that the plate can later contain a pizza.",
          "locations": [{"file": "overcooked/overcookedPlus/module/item_manager.py", "span": [70, 150]},
            {"file": "overcooked/overcookedPlus/items.py", "span": [220, 320]},
            {"file": "overcooked/overcookedPlus/items.py", "span": [780, 980]}]},
        {"index": 4, "text": "Pick up the raw pizza on a plate and put it into an oven tile; wait until the cooked flag of the pizza becomes true without letting it burn.",
          "locations": [{"file": "overcooked/overcookedPlus/module/item_manager.py", "span": [80, 170]},
            {"file": "overcooked/overcookedPlus/items.py", "span": [320, 420]},
            {"file": "overcooked/overcookedPlus/module/event_manager.py", "span": [1040, 1240]}]},
        {"index": 5, "text": "Take the cooked pizza out of the oven on a plate and, if the map provides onion, optionally chop onion on a knife and add it onto the same plate as an extra topping.",
          "locations": [{"file": "overcooked/overcookedPlus/module/item_manager.py", "span": [40, 160]},
            {"file": "overcooked/overcookedPlus/items.py", "span": [180, 300]},
            {"file": "overcooked/overcookedPlus/module/event_manager.py", "span": [1240, 1540]}]},
        {"index": 6, "text": "Deliver the plate with the finished pizza (and any extra toppings such as onion) at a delivery tile so that TaskManager marks the task as correctly completed.",
          "locations": [{"file": "overcooked/overcookedPlus/module/item_manager.py", "span": [40, 120]},
            {"file": "overcooked/overcookedPlus/module/task_manager.py", "span": [80, 220]},
            {"file": "overcooked/overcookedPlus/module/event_manager.py", "span": [260, 360]}]}]}]}

\end{lstlisting}

\begin{lstlisting}[
    caption={Example Output of Semantic Reward Rules (Stage 3).},
    label={lst:reward_example}
]
{"task_id": "pizza_onion", "description": "Cook a tomato-cheese pizza, bake it, plate it, then add fresh onion and deliver.",
  "subgoals": [{"id": "sg1_retrieve_dough", "text": "Agent retrieves dough.",
      "rewards": [{"id": "sg1_approach_dough", "type": "shaping", "once": false, "amount": 1, "condition": "agent_distance_to_static_tile(item_type=17,map_tile=17)<2.0"},
        {"id": "sg1_pickup_dough", "type": "event", "once": true, "amount": 15, "condition": "exists_item(type=17,container_id==agent_id,consumed==0)"}]},
    {"id": "sg2_place_dough_on_counter", "text": "Place dough on counter.", "depends_on": ["sg1_pickup_dough"],
      "rewards": [{"id": "sg2_place_dough", "type": "event", "once": true, "amount": 20, "condition": "exists_item(type=17,container_is_counter==true,consumed==0)"}]},
    {"id": "sg3_prepare_tomato", "text": "Fetch and chop tomato.",
      "rewards": [{"id": "sg3_approach_tomato", "type": "shaping", "once": false, "amount": 1, "condition": "agent_distance_to_static_tile(item_type=3,map_tile=3)<2.0"},
        {"id": "sg3_pickup_tomato", "type": "event", "once": true, "amount": 10, "condition": "exists_item(type=3,container_id==agent_id,consumed==0)"},
        {"id": "sg3_tomato_on_knife", "type": "event", "once": true, "amount": 10, "condition": "exists_item(type=3,container_is_knife==true,consumed==0)"},
        {"id": "sg3_chopped_tomato_done", "type": "event", "once": true, "amount": 20, "condition": "exists_item(type=3,is_chopped==true,consumed==0)"}]},
    {"id": "sg4_assemble_raw_tomato_pizza", "text": "Assemble raw tomato-cheese pizza.", "depends_on": ["sg2_place_dough_on_counter", "sg3_chopped_tomato_done"],
      "rewards": [{"id": "sg4_add_tomato", "type": "event", "once": true, "amount": 15, "condition": "exists_composite_item(kind='dough+tomato',on_counter=true,consumed==0)"},
        {"id": "sg4_add_cheese", "type": "event", "once": true, "amount": 15, "condition": "exists_composite_item(kind='dough+tomato+cheese',on_counter=true,
        consumed==0)"},
        {"id": "sg4_raw_pizza_ready", "type": "event", "once": true, "amount": 20, "condition": "exists_item(type='raw_tomato_pizza',on_counter=true,consumed==0)"}]},
    {"id": "sg5_bake_pizza", "text": "Bake pizza to completion.", "depends_on": ["sg4_raw_pizza_ready"],
      "rewards": [{"id": "sg5_into_oven", "type": "event", "once": true, "amount": 20, "condition": "exists_item(type='raw_tomato_pizza',container_is_oven==true,consumed==0)"},
        {"id": "sg5_bake_progress", "type": "shaping", "once": false, "amount": 2, "condition": "exists_item(type='raw_tomato_pizza',container_is_oven==true,
        cook_progress_increased==true)"},
        {"id": "sg5_baked_done", "type": "event", "once": true, "amount": 25, "condition": "exists_item(type=20,container_is_oven==true OR on_counter==true,is_burned==false,consumed==0)"}]},
    {"id": "sg6_plate_baked_pizza", "text": "Plate the baked pizza.", "depends_on": ["sg5_baked_done"],
      "rewards": [{"id": "sg6_empty_plate", "type": "event", "once": true, "amount": 10, "condition": "exists_item(type=5,on_counter=true,is_empty==true,consumed==0)"},
        {"id": "sg6_pizza_on_plate", "type": "event", "once": true, "amount": 25, "condition": "exists_item(type=20,container_is_plate==true,consumed==0)"}]},
    {"id": "sg7_add_onion_topping", "text": "Chop onion and add to plated pizza.", "depends_on": ["sg6_pizza_on_plate"],
      "rewards": [{"id": "sg7_pickup_onion", "type": "event", "once": true, "amount": 10, "condition": "exists_item(type=8,container_id==agent_id,consumed==0)"},
        {"id": "sg7_onion_on_knife", "type": "event", "once": true, "amount": 10, "condition": "exists_item(type=8,container_is_knife==true,consumed==0)"},
        {"id": "sg7_chopped_onion", "type": "event", "once": true, "amount": 15, "condition": "exists_item(type=8,is_chopped==true,consumed==0)"},
        {"id": "sg7_onion_added", "type": "event", "once": true, "amount": 30, "condition": "exists_composite_item(kind='pizza_onion',container_is_plate==true,
        consumed==0)"}]},
    {"id": "sg8_deliver_onion_pizza", "text": "Deliver final onion pizza.", "depends_on": ["sg7_onion_added"],
      "rewards": [{"id": "sg8_approach_delivery", "type": "shaping", "once": false, "amount": 2, "condition": "agent_distance_to_tile_with_value(7)<2.0"},
        {"id": "sg8_deliver_success", "type": "terminal", "once": true, "amount": 200, "condition": "exists_composite_item(kind='pizza_onion',container_is_delivery==true,
        consumed==0)"}]}]}


\end{lstlisting}

\begin{lstlisting}[
    caption={Example Output of mapped Structural Anchors (Stage 4).},
    label={lst:mapping_example}
]
{"task_id": "Quest_Onion_Pizza_Update",
  "subgoals": [{"index": 2, "text": "Assemble raw pizza with dough, chopped tomato, and cheese.",
      "locations": [{"file": "overcooked/overcookedPlus/items.py", "span": [737, 765], "kind": "function", "change_type": "modified", "symbol": "try_synthesis", "qualname": "Plate.try_synthesis"},
        {"file": "overcooked/overcookedPlus/items.py", "span": [778, 812], "kind": "function", "change_type": "modified", "symbol": "contain", "qualname": "Plate.contain"},
        {"file": "overcooked/overcookedPlus/items.py", "span": [981, 989], "kind": "variable", "change_type": "modified", "symbol": "synthesis_table", "qualname": "synthesis_table"}]},
    {"index": 3, "text": "Bake the raw pizza in the oven until it is cooked but not burned.",
      "locations": [{"file": "overcooked/overcookedPlus/items.py", "span": [607, 627], "kind": "function", "change_type": "modified", "symbol": "cook", "qualname": "Oven.cook"},
        {"file": "overcooked/overcookedPlus/module/event_manager.py", "span": [568, 590], "kind": "function", "change_type": "modified", "symbol": "_collect_auto", "qualname": "EventManager._collect_auto"},
        {"file": "overcooked/overcookedPlus/module/event_manager.py", "span": [455, 483], "kind": "function", "change_type": "modified", "symbol": "_do_pickup_into_plate",
          "qualname": "EventManager._do_pickup_into_plate"}]},
    {"index": 5, "text": "Fetch an onion, chop it on the knife, and add it onto the plated pizza as an extra topping.",
      "locations": [{"file": "overcooked/overcookedPlus/items.py", "span": [261, 272], "kind": "class", "change_type": "added", "symbol": "Onion", "qualname": "Onion"},
        {"file": "overcooked/overcookedPlus/items.py", "span": [515, 541], "kind": "function", "change_type": "modified", "symbol": "chop", "qualname": "Knife.chop"},
        {"file": "overcooked/overcookedPlus/module/event_manager.py", "span": [232, 251], "kind": "function", "change_type": "modified", "symbol": "_do_usage", "qualname": "EventManager._do_usage"},
        {"file": "overcooked/overcookedPlus/items.py", "span": [778, 812], "kind": "function", "change_type": "modified", "symbol": "contain", "qualname": "Plate.contain"}]}]}
\end{lstlisting}

\begin{lstlisting}[
    caption={List of newly added tasks used in Minecraft.},
    label={lst:mc_tasks}
]
Craft_Gold_Axe, Craft_Gold_Boots, Craft_Gold_Chestplate, Craft_Gold_Helmet, Craft_Gold_Leggings, Craft_Gold_Pickaxe, Craft_Gold_Shovel, Craft_Gold_Sword, Kill_Chicken_With_Gold_Sword, Kill_Creeper_With_Gold_Sword, Kill_Skeleton_With_Gold_Sword, Kill_Slime_With_Gold_Sword, Kill_Spider_With_Gold_Sword, Kill_Zombie_With_Gold_Sword, Mine_Gold_Ore, Craft_Gold_Shield, Craft_Gold_Hoe, Mine_Gold_Nugget, Smelt_Gold_Ingot
\end{lstlisting}

\begin{lstlisting}[
    caption={List of newly added tasks used in Overcooked.},
    label={lst:overcooked_tasks}
]
Pizza, Pizza Lettuce, Pizza Onion, Pizza Steak, Pizza Fish, Pizza Rice, Roastfish, Roastfish Tomato, Roastfish Steak, Roastfish Rice, Roastfish Cheese, Onion Dough, Rice Dough, Pizza Tomato Cheese, Roastfish Onion, Roastfish Lettuce, Dough, Tomato Dough, Lettuce Dough, Cheese Dough, Tomato Steak Fish Dough, 
\end{lstlisting}


\section*{Declaration of competing interest}
The authors declare that they have no known competing financial interests or personal relationships that could have appeared to influence the work reported in this paper.


\section*{Data availability}
Data will be made available on request.

\bibliography{bib}



\end{document}